\documentclass{article}

\usepackage{amsmath,amsfonts,bm}

\def\eqref#1{equation~\ref{#1}}

\def\1{\bm{1}}

\def\vy{{\bm{y}}}

\def\mF{{\bm{F}}}

\def\mI{{\bm{I}}}

\def\mP{{\bm{P}}}

\def\mR{{\bm{R}}}

\DeclareMathAlphabet{\mathsfit}{\encodingdefault}{\sfdefault}{m}{sl}
\SetMathAlphabet{\mathsfit}{bold}{\encodingdefault}{\sfdefault}{bx}{n}

\usepackage{microtype}
\usepackage{graphicx}
\usepackage{subfigure}
\usepackage{booktabs} 

\usepackage{hyperref}
\usepackage{url}
\usepackage[utf8]{inputenc} 
\usepackage[T1]{fontenc}    
\usepackage{booktabs}       
\usepackage{amsfonts}       
\usepackage{nicefrac}       
\usepackage{microtype}      
\usepackage{xcolor}         
\usepackage{graphicx}
\usepackage{natbib}
\usepackage{amsmath}
\usepackage{paralist}
\usepackage{multirow}
\usepackage{caption}
\usepackage{subcaption}
\usepackage{xcolor}
\usepackage{wrapfig}

\usepackage{hyperref}

\usepackage[accepted]{icml2024}

\usepackage{amsmath}
\usepackage{amssymb}
\usepackage{mathtools}
\usepackage{amsthm}
\usepackage{hyperref}
\usepackage{balance}

\usepackage[capitalize,noabbrev]{cleveref}

\theoremstyle{plain}

\theoremstyle{definition}

\theoremstyle{remark}

\usepackage[textsize=tiny]{todonotes}

\icmltitlerunning{Revealing Vision-Language Integration in the Brain with Multimodal Networks}

\begin{document}

\twocolumn[
\icmltitle{Revealing Vision-Language Integration in the Brain with Multimodal Networks}


\icmlsetsymbol{equal}{*}

\begin{icmlauthorlist}
\icmlauthor{Vighnesh Subramaniam}{yyy,comp}
\icmlauthor{Colin Conwell}{sch}
\icmlauthor{Christopher Wang}{yyy,comp}
\icmlauthor{Gabriel Kreiman}{sch1}
\icmlauthor{Boris Katz}{yyy,comp}
\icmlauthor{Ignacio Cases}{yyy,comp}
\icmlauthor{Andrei Barbu}{yyy,comp}
\end{icmlauthorlist}

\icmlaffiliation{yyy}{MIT CSAIL}
\icmlaffiliation{comp}{CBMM}
\icmlaffiliation{sch}{Department of Cognitive Science, Johns Hopkins University}
\icmlaffiliation{sch1}{Boston Children’s Hospital, Harvard Medical School}

\icmlcorrespondingauthor{Vighnesh Subramaniam}{vsub851@mit.edu}

\icmlkeywords{Machine Learning, ICML}

\vskip 0.3in
]

\printAffiliationsAndNotice{}  

\begin{abstract}
We use (multi)modal deep neural networks (DNNs) to probe for sites of multimodal integration in the human brain by predicting stereoencephalography (SEEG) recordings taken while human subjects watched movies.  We operationalize sites of multimodal integration as regions where a multimodal vision-language model predicts recordings better than unimodal language, unimodal vision, or linearly-integrated language-vision models. Our target DNN models span different architectures (e.g., convolutional networks and transformers) and multimodal training techniques (e.g., cross-attention and contrastive learning). As a key enabling step, we first demonstrate that trained vision and language models systematically outperform their randomly initialized counterparts in their ability to predict SEEG signals. We then compare unimodal and multimodal models against one another. Because our target DNN models often have different architectures, number of parameters, and training sets (possibly obscuring those differences attributable to integration), we carry out a controlled comparison of two models (SLIP and SimCLR), which keep all of these attributes the same aside from input modality. Using this approach, we identify a sizable number of neural sites (on average 141 out of 1090 total sites or 12.94\%) and brain regions where multimodal integration seems to occur. Additionally, we find that among the variants of multimodal training techniques we assess, CLIP-style training is the best suited for downstream prediction of the neural activity in these sites.
\end{abstract}

\section{Introduction}
\label{sec:introduction}

We expand the use of deep neural networks for understanding the brain from unimodal models, which can be used to investigate language and vision regions in isolation, to multimodal models, which can be used to investigate vision-language integration. Beginning with work in the primate ventral visual stream \citep{yamins2014performance, schrimpf2020brain}, this practice now includes the study of both the human vision and language cortex alike \citep{chang2019bold5000,allen2021massive, bhattasali2020alice, nastase2021narratives,schrimpf2021neural, goldstein2021thinking, goldstein2022shared, lindsay2021convolutional, caucheteux2022brains, conwell2023pressures}. These studies, however, focus on a single modality of input --- vision alone or language alone. Yet, much of what humans do fundamentally requires multimodal integration.

As a product of the unimodal focus, we have learned far less about the correspondence between biological and artificial neural systems tasked with processing visual and linguistic input \textit{simultaneously}. Here, we seek to address this gap by using performant, pretrained multimodal deep neural network (DNN) models (ALBEF \citep{li2021align}, BLIP \citep{li2022blip}, Flava \citep{singh2022flava}, SBERT \citep{reimers-2019-sbert}, BEIT \citep{bao2021beit}, SimCSE \citep{gao2021simcse}, SIMCLR \citep{chen2020simple}, CLIP \citep{radford2021learning}, SLIP \citep{mu2021slip}) to predict neural activity in a large-scale stereoelectroencephalography (SEEG) dataset consisting of neural responses (from intracranial electrodes) to the images (frames) and dialog of popular movies \citep{yaari2022aligned}. Our primary goal in this work is to use systematic comparisons between the neural predictivity of unimodal and multimodal models to probe for sites of vision-language integration in the brain. 

Our work make the following contributions:
\begin{compactenum}
\item We introduce a rigorous, multi-stage statistical analysis to compare multimodal models against neural data, against one another, and against unimodal models. In tandem, we release a code toolbox to perform this analysis and enable future work.
\item We demonstrate that this approach is sufficiently robust to distinguish randomly initialized from trained vision, language, and multimodal models. Previous work, especially in the prediction of language-evoked brain activity, has occasionally found little difference between randomly-initialized and trained models, and there has been no systematic review of this difference with respect to multimodal architectures. (Without establishing this difference, we could not conclude that multimodal processing is taking place, only that multimodal architectures are generically helpful.)
\item Using a wide array of models, we deploy our modeling procedure to identify areas associated with multimodal processing, i.e. areas where multimodal models outperform unimodal models as well as language-vision models with linearly-integrated features, e.g., concatenation of vision and language features.
\item We then introduce an architecture-, parameter-, and dataset-controlled experiment where two variants of the same model, one unimodal and the other multimodal, are used to identify multimodal brain regions.
\item We catalogue a collection of brain areas (and individual neural sites) that appear to benefit from multimodal integration, and further specify \textit{which} multimodal models best explain the activity in areas associated with this integration.
\end{compactenum}

\begin{figure*}
\centering
    \includegraphics[width = 0.8\textwidth]{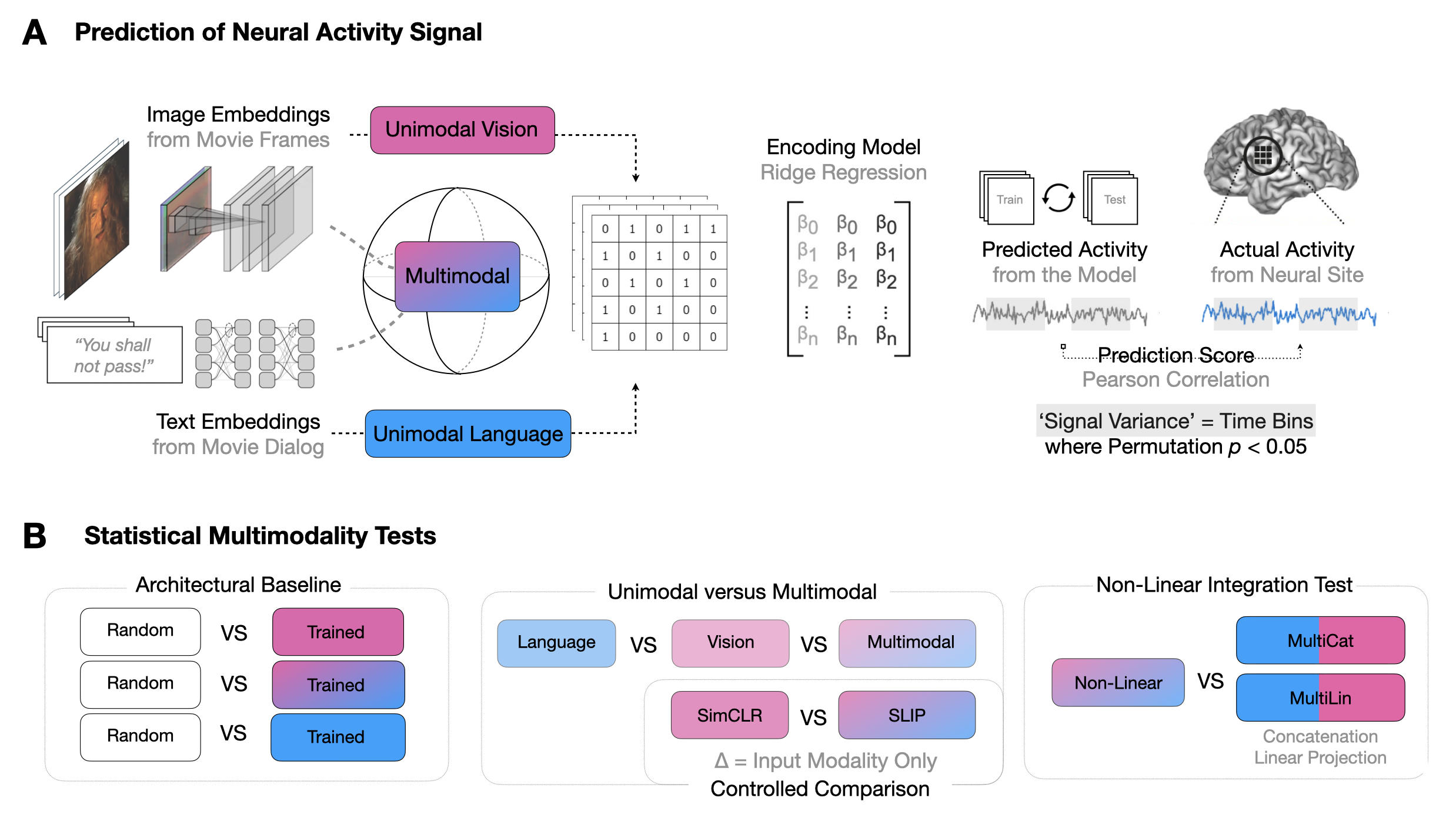}
    \vspace{-2ex}
    \caption{\textbf{Overview}. (A) We parse the stimuli, movies, into image-text pairs (which we call \emph{event structures}) and process these with either a vision model, text model, or multimodal model. We extract feature vectors from these models and predict neural activity in 161 25ms time bins per electrode, obtaining a Pearson correlation coefficient per time bin per electrode per model. We exclude any time bins in which a bootstrapping test (computed over event structures) suggests an absence of meaningful signal in the neural activity target in that bin. We run these regressions using both trained and randomly initialized encoders and for two datasets, a vision-aligned dataset and language-aligned dataset, which differ in the methods to sample these pairs. (B) The first analysis of this data investigates if trained models outperform randomly initialized models. The second analysis investigates if multimodal models outperform unimodal models. The third analysis repeats the second holding constant the architecture and dataset to factor out these confounds. A final analysis investigates if multimodal models that meaningfully integrate vision and language features outperform models that simply concatenate them.}
    \label{fig:methods}
    \vspace{-1ex}
\end{figure*}

Taken together, these experiments more directly connect multimodal networks and multimodal regions in the brain. In so doing, they also allow us to assess which current neural networks of vision and language are best suited for the modeling of brain-like multimodal integration.\footnote{Code, data, and annotations to reproduce our results are available at {\scriptsize\url{github.com/vsubramaniam851/brain-multimodal/}}}. We give an overview of our approach in Figure~\ref{fig:methods}.

\section{Related Work}

\textbf{Multimodal Processing}. Multimodal vision and language processing in the brain is presumed to show some degree of localization based on neuroscience experiments where subjects are presented with specially constructed multimodal visio-linguistic stimuli and the response is measured using functional Magnetic Resonance Imaging (fMRI) against several control stimuli. 

For example, recent multivariate fMRI studies have identified the superior temporal cortex to be associated with specific forms of auditory and visual processing \citep{jouen2015beyond, zhang2023multivariate, van2023exploring, friederici2009role, friederici2012cortical}.  Furthermore, deeper semantic integration of vision and language has been seen in the middle temporal and inferior parietal cortex \citep{petrides2023evolution, bzdok2016left}. Other areas include the supramarginal gyrus, involved in emotion processing \citep{stoeckel2009supramarginal}, the superior frontal lobe, commonly associated with self-awareness \citep{schilling2013cortical}, the caudal middle frontal cortex, commonly associated with eye movements and scene understanding \citep{badre2018frontal}, and the pars orbitalis, which contains Broca's area and is associated with speech processing \citep{belyk2017convergence}. 

\textbf{DNN Prediction of Brain Activity}. There has been considerable interest in investigating the emergent alignment of deep neural network (DNN) representation to representation in the biological brain \citep{wehbe2014aligning, kuzovkin2018activations, conwell2021neural, goldstein2021thinking, goldstein2022shared, lindsay2021convolutional, caucheteux2022brains}. These approaches have typically employed various forms of linear regressions to predict biological brain activity from the internal states of candidate models, with specific modifications to the prediction pipeline depending on the type of neural recording used. 

The majority of these works tend to focus on vision or language alone, in large part because unimodal datasets \citep{chang2019bold5000,allen2021massive, bhattasali2020alice, nastase2021narratives} and unimodal models (e.g., PyTorch-Image-Models; Hugging Face) are the most commonly available. Many of these works have shown that language-based unimodal networks and vision-based unimodal networks are highly predictive of the brain activity evoked by inputs in their respective modalities. Some of these works also include comparisons between trained and randomly initialized networks and have shown that trained unimodal vision networks model activity better than randomly initialized networks, but that a similar effect is not always evident in the modeling of language \citep{schrimpf2021neural}.

More recent work has suggested that multimodal models can in certain cases predict image-evoked fMRI activity in human high-level visual cortex better than unimodal vision models \citep{wang2023better, oota2022visio}. Our work differs from this and other prior work by considering general vision-language integration in a dataset defined by simultaneous inputs from vision and language alike (i.e. movies). We employ multimodal networks including ALBEF, CLIP, and SLIP and use representations from these networks to predict brain activity up to $2000$ms before and after the occurrence of an event. Our results unveil a number of distinct electrodes wherein the activity from multimodal networks predicts activity better than any unimodal network, in ways that control for differences in architecture, training dataset, and integration style where possible. In contrast to most previous work that mainly leverages fMRI, here we focus on high-fidelity neurophysiological signals \citep{kuzovkin2018activations}. Our analysis unveils a number of candidate sites for multimodal integration (many of which align with and overlap with areas mentioned in prior work) and allows us to rank the kinds of multimodal integration (in the models) that best predict the activity at these sites.

\section{Methods}
\label{sec:methods}

\textbf{Neural Data}: Invasive intracranial field potential recordings were collected during $7$ sessions from $7$ subjects ($4$ male, $3$ female; aged $4 - 19$, $\mu = 11.6$, $\sigma = 4.6$) with pharmacologically intractable epilepsy. During each session, subjects watched a feature-length movie from the Aligned Multimodal Movie Treebank (AMMT) \citep{yaari2022aligned} in a quiet room while neural activity was recorded using SEEG electrodes \citep{liu2009timing} at a rate of $2$kHz.

We parse the neural activity and movie into language-aligned events (word onset) and visually-aligned events (scene cuts) where each event consists of an individual image-text pair and create two stimulus alignment datasets where we have coregistered visual and language inputs to the given models. Each element of the dataset consists of a stimulus and the corresponding brain activity. For the language-aligned dataset, the stimulus data are word utterances with their sentence context as well as the corresponding closest movie frame to the word onset. For the vision-aligned dataset, the stimulus data are the frames from eachscene cut and the closest sentence to occur after the cut.

Word-onset times are collected as part of the AMMT metadata and visual scene cuts are extracted from each movie using PySceneDetect \citep{pyscenedetect}. For the corresponding brain activity, following \citet{goldstein2021thinking}, we extract a $4000$ms window of activity (about $8000$ samples), $2000$ms prior to the event occurrence and $2000$ms after the event occurrence, per electrode. We split the $4000$ms window into sub-windows of $200$ms with a sliding window of $25$ms. We then average the activity per sub-window to get a series of mean activity values over time per electrode. Further details of the neural data processing can be found in Appendix~\ref{appendix:neural-event-structures}.

\textbf{Models}: We examine $12$ pretrained deep neural network models, $7$ multimodal and $5$ unimodal, to explore the effect of multimodality on predictions of neural activity. To identify areas associated with \emph{multimodal integration}, we choose multimodal models that directly integrate vision and language. Mathematically, this means that we choose models that apply non-linear transformations on vision and language features directly in model computations either through (i) cross-attention or (ii) contrastive training. 

The models that serve as our most controlled experimental contrast are the SLIP models \citep{mu2021slip}. The SLIP models are a series of 3 models that use the same architecture (ViT-[S,B,L]) and the same training dataset (YFCC15M), but are trained with one of three objective functions: (1) pure unimodal SimCLR-style \citep{chen2020simple} visual contrastive learning (henceforth SLIP-SimCLR), (2) pure multimodal CLIP-style \citep{radford2021learning} vision-language alignment (henceforth SLIP-CLIP), and (3) combined visual contrastive learning with multimodal CLIP-style vision-language alignment (henceforth SLIP-Combo). The full set constitutes a set of 5 models (SLIP-SimCLR; the SLIP-CLIP visual encoder; the SLIP-CLIP language encoder; the SLIP-Combo visual encoder; the SLIP-Combo language encoder).

For more general (uncontrolled) multimodal-unimodal contrasts, we first include architecturally multimodal networks ALBEF \citep{li2021align}, BLIP \citep{li2022blip}, and Flava \citep{singh2022flava}. These are networks that use non-linear cross-attention mechanisms to integrate vision and language instead of using contrastive training, and therefore, vision-language integration is directly incorporated into the design of the network computations. We also include unimodal models SBERT, \citep{reimers2019sentence}, SimCSE \citep{gao2021simcse}, BEIT \citep{bao2021beit}, and ConvNeXt \citep{liu2022convnet}. Additionally, we design two \emph{linearly-integrated language-vision networks} of our own. Our first model, \emph{MultiConcat}, concatenates representations from a pretrained SimCSE and pretrained SLIP-SimCLR. Our second model, \emph{MultiLin}, performs the same concatenation and trains a linear projection using the NLVR-2 \citep{suhr2018corpus} dataset. 

For each of the $14$ networks, we run experiments on both pretrained and randomly-initialized weights to assess whether the multimodality we assume in the brain coincides with features learned in training the multimodal models. Random initialization of these networks has different effects on the multimodal status of particular networks. Since SLIP-Combo and SLIP-CLIP are designed to be multimodal due to contrastive training, randomly initialized SLIP-Combo or SLIP-CLIP are considered unimodal. (The multimodal signal used to guide model predictions is lost due to the random initialization in this case.). ALBEF, BLIP, and Flava, on the other hand, are architecturally multimodal models that directly take both modalities as input regardless of random initialization. Random initialization for these three networks has no effect on the multimodal status of output representations. Details on the reasoning behind these choices are given in Appendix~\ref{appendix:networks}.

\textbf{Neural Regression}: To identify multimodal electrodes and regions in the brain, we first extract feature vectors from every layer of the candidate networks using the image-text pairs in a given dataset alignment. We then use these features from each layer as predictors in a $5$-fold ridge regression predicting the averaged neural activity of a target neural site in response to each \emph{event structure} (defined here as an image-text pair). Per fold, we split our dataset of event structures contiguously based on occurrence in the movie. We place $80\%$ of the event structures in the training set, $10\%$ of event structures in the validation set, and $10\%$ in the testing set. We use contiguous splitting to control for the autoregressive nature of the movie stimuli. We measure the strength of the regression using the Pearson correlation coefficient between predicted average activity and actual average activity for a \textit{specific} time window in each neural site for a held-out test set of event structures. 

Two aspects of this process are worth emphasizing: First, the final performance metric (the Pearson correlation between actual and predicted neural activity for a held-out test set of event-structures) is not a correlation over time-series (for which the Pearson correlation is inappropriate), but a correlation over a set of (nominally IID) event-structures that we have extracted by design to minimize the autoregressive confounds of time-series data. Second, the cross-validation procedure and train-test splitting is specifically designed to assess the generalization of the neural regression fits, and as such contains no cross-contamination of selection procedures (e.g., the maximally predictive layer from a candidate model, feature normalization, or the ridge regression lambda parameter) and final model scoring. In this case, we use the cross-validation procedure to select the scores associated with the best performing layer and select the best performing regression hyperparameters. Further details on the regression method can be seen in Appendix \ref{appendix:regression}.

\textbf{Bootstrapped Confidence Intervals Across Time}: In order to make model comparisons on a sound statistical basis, we use a bootstrapping procedure over the image-text pairs in a given dataset alignment to calculate $95\%$ confidence intervals on the correlation scores per time bin for the training, validation, and test set alike.

Our bootstrapping procedure involves first resampling the image-text pairs and corresponding neural activity with replacement and then re-running the regression with the resampled event structures, predicting the associated neural activity per time bin per electrode. We run the resampling $1000$ times and use the same resampled event structures across all models to allow for model comparison. Directly mimicking the standard encoding procedure, this bootstrapping leaves us with 95\% confidence intervals on the predictive accuracy of a given model per time bin per electrode across all of the training, validation, and test splits. We obtain two sets of confidence intervals per dataset alignment, either language- or vision-aligned. In subsequent model comparisons, we use the 95\% confidence interval over the validation set to filter out time bins per electrode in which either of the model's scores was not significantly above 0. Subsequent analysis uses the held-out test set scores for analysis.

\textbf{Model Comparisons}:
Taking inspiration from fMRI searchlight analyses \citep{kriegeskorte2006information, etzel2013searchlight}, we next perform a series of statistical tests on each electrode to determine whether or not they are better predicted by multimodal or unimodal representations and whether each electrode is better predicted by representations from trained models or randomly initialized models.

We first filter all time bins in electrodes for models where the lower $95\%$ confidence interval of the validation score overlaps with zero. This ensures that the analysis focuses on time bins and electrodes with meaningful neural signal. We remove models from further analysis on a particular electrode if that model has a confidence interval that overlaps with zero for all time bins on the validation set. If only one model has at least $10$ time bins with this requirement (a minimal threshold for bootstrapped comparisons), we consider this model the best model by default and do no further processing on the electrode. 

For electrodes without these ``default winners'', we employ an additional statistical test of model difference between the first and second-highest ranking models for a given comparison. That is, we use a second-order bootstrapping procedure (this time across time bins, rather than across event structures), calculating the difference in the average score across resampled time bins between the 2 candidate models in a given comparison. This procedure minimizes the possibility of one model producing a random peak of predictivity that does not adequately reflect its predictivity more generally, and may artificially give the impression of being the superior model in a comparison. We run this for model pairs on electrodes that have at least $10$ time bins remaining after filtering based on the lower confidence interval of the validation set for both models. For the bootstrapping procedure of model difference, we identify electrodes where the difference in performance is statistically significant and use FDR (Benjamni-Hochberg) multiple comparisons correction to adjust the p-value (per electrode) on each test.

\textbf{Multimodality Tests}: The multimodality logic we apply (in order of stringency) is as follows: (1) \textbf{Weak test of multimodality}: Is ANY multimodal model or linearly-integrated vision-language model significantly more predictive than all other unimodal models in \textit{either} of our dataset alignments (word onset, scene cuts)? (2) \textbf{Weak SLIP test}: Is the SLIP-Combo vision transformer significantly more predictive than the SLIP-SimCLR vision transformer in \textit{either} of our dataset alignments? (3) \textbf{Strict test of multimodality}: Is ANY multimodal model or linearly-integrated vision-language model significantly more predictive than all other unimodal models in BOTH of our dataset alignments? (4) \textbf{Strict SLIP test}: Is the SLIP-Combo vision transformer more predictive than SLIP-SimCLR vision transformer in BOTH of our alignments? (5) \textbf{Non-linear integration test}: For electrodes that pass test 3, is a multimodal model more predictive than both linearly-integrated vision-language models, i.e. MultiConcat and MultiLin? (A more detailed description of these tests is given in Appendix~\ref{appendix:multi-test}).

For these tests, we use both the ``default winner analysis'' (i.e. an electrode passing automatically if the only model left after filtering is either multimodal or SLIP-Combo more specifically), and the bootstrapped model comparison test. Tests 2 and 4 control for architecture and dataset, which ensures that models cannot be outperformed due to architecture, hyper-parameters, or the training dataset. For all electrodes that pass our multimodality test, we use our model comparisons to identify the best multimodal architecture for explaining activity in the electrode.
\vspace{-2ex}

\section{Results}
While there is no single meaningful measure of overall modeling performance, since we expect significant variance in performance as a function of \textit{multiple} controlled and uncontrolled sources, there are a few key metrics we can consider to provide an overall gestalt of our model-to-brain encoding pipeline and the specific measured effects. Unless otherwise noted, we use the following convention in the reporting of these metrics: arithmetic average over the bootstrapped mean scores [lower 95\% confidence interval; upper 95\% confidence interval].

As an initial heuristic, we consider the bootstrapped test set score mean, as well as the bootstrapped test mean upper and lower bounds on performance across all N = 28 models (14 architectures, with both trained and randomly-initialized weights), N = 2 dataset alignments (word onsets, scene cuts) and all N = 1090 electrodes, after we've selected the max accuracy across time. This constitutes a total of 24 * 2 * 1090 = 39,420 data points.

The bootstrapped global average (i.e. an average across the bootstrapped means) across these data points is $r_{\text{Pearson}}$ = 0.142 [0.0797, 0.269]. The bootstrapped maximum across the bootstrapped means is $r_{\text{Pearson}}$ = 0.539 [0.517, 0.561]. And the bootstrapped minimum across the bootstrapped means is $r_{\text{Pearson}}$ =  -0.223  [-0.398, -0.034]. (Negatives here mean model predictions were anticorrelated with ground truth.) This is of course a coarse metric, meant only to give some sense of the encoding performance overall, and to demonstrate its notable range across electrodes. (For an overview of results, see Figure~\ref{fig:new_results}, which shows our average model performance across several brain regions of interest.)

\subsection{Trained versus Randomly Initialized Results}
\begin{figure*}
\centering
    \includegraphics[width=0.84\textwidth]{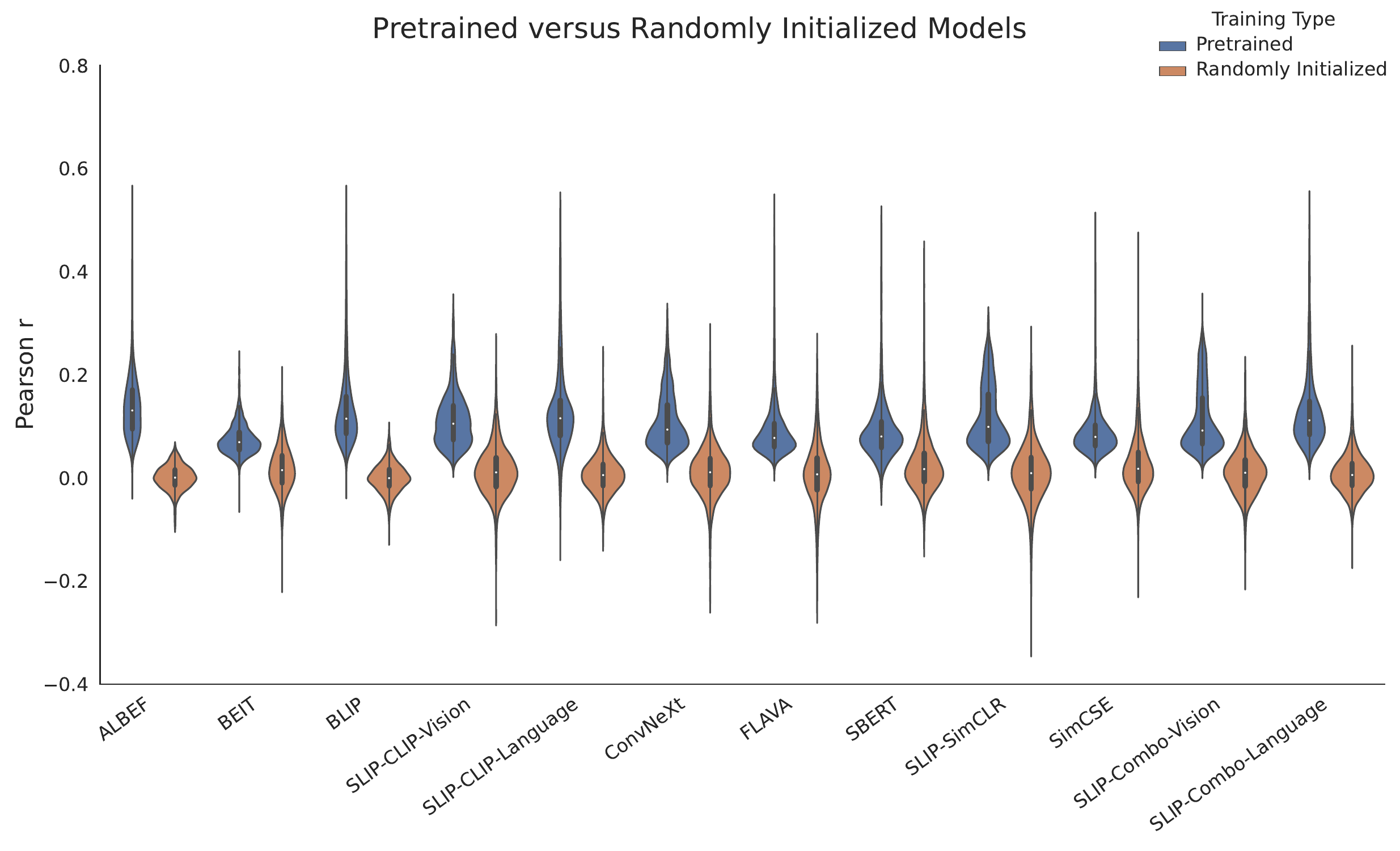}
    \caption{\textbf{Trained models beat randomly initialized models.} A comparison between pretrained and randomly initialized model performance showing the distribution of predictivity across electrodes. This averages significant time bins per electrode (where the lower validation confidence interval must be greater than zero), for both datasets alignments and for each of our 12 models. Every trained network outperforms its randomly initialized counterpart. Trained networks overall outperform untrained networks. This is true both on average, and for almost every single electrode.}
    \label{fig:train_random}
\end{figure*}
 We first use the comparison methods to analyze the difference between neural predictivity of trained models and neural predictivity of randomly initialized models. After filtering out models and time bins in electrodes where the lower validation confidence interval score is less than zero, this leaves us with 498/1090 unique electrodes. We show the average difference in performance for these electrodes in Figure~\ref{fig:train_random}. In 120 of these electrodes, the default model was a trained model after filtering according to the default winners analysis. For the rest of the 278 electrodes, we use a bootstrapping comparison on the remaining electrodes assuming models have at least 10 time bins remaining.
 
 We find that trained models beat randomly initialized models on all 278 electrodes according to the bootstrapping comparison. The average difference in scores across the dataset alignments was $r_{\text{Pearson}} = 0.107 [0.026, 0.238]$, showing the significant improvement that trained models have over randomly initialized models. These results demonstrate that experience and structured representations are necessary to predict neural activity in our case for any network, regardless of whether the network is a language network, visual network, or multimodal network.

\subsection{Multimodality Test Results}
\label{sec:mul-tests}

\begin{table}
\resizebox{\columnwidth}{!}{%
\begin{tabular}{lcc}
\toprule
\textbf{Test}                         & \textbf{Language-Aligned Electrodes} & \textbf{Vision-Aligned Electrodes} \\ \midrule
Weak test of multimodality   & 213                    & 60                   \\
Weak SLIP test               & 218                    & 73                   \\
Strict test of multimodality & 12                     & 12                   \\
Strict SLIP test             & 28                     & 28                  \\
Non-linear integration test  & 28                     & 28                   \\ \bottomrule
\end{tabular}
}
\caption{\textbf{Multimodal Integration Tests}. We design five multimodal integration tests of varying strictness to test each electrode for multimodal integration. We report the number of electrodes out of 1090 total electrodes that pass each specific test for our language-aligned (left) and vision-aligned (right) datasets.}
\label{tab:mul-tests}
\end{table}

\begin{figure*}
    \centering
    \includegraphics[width=\textwidth]{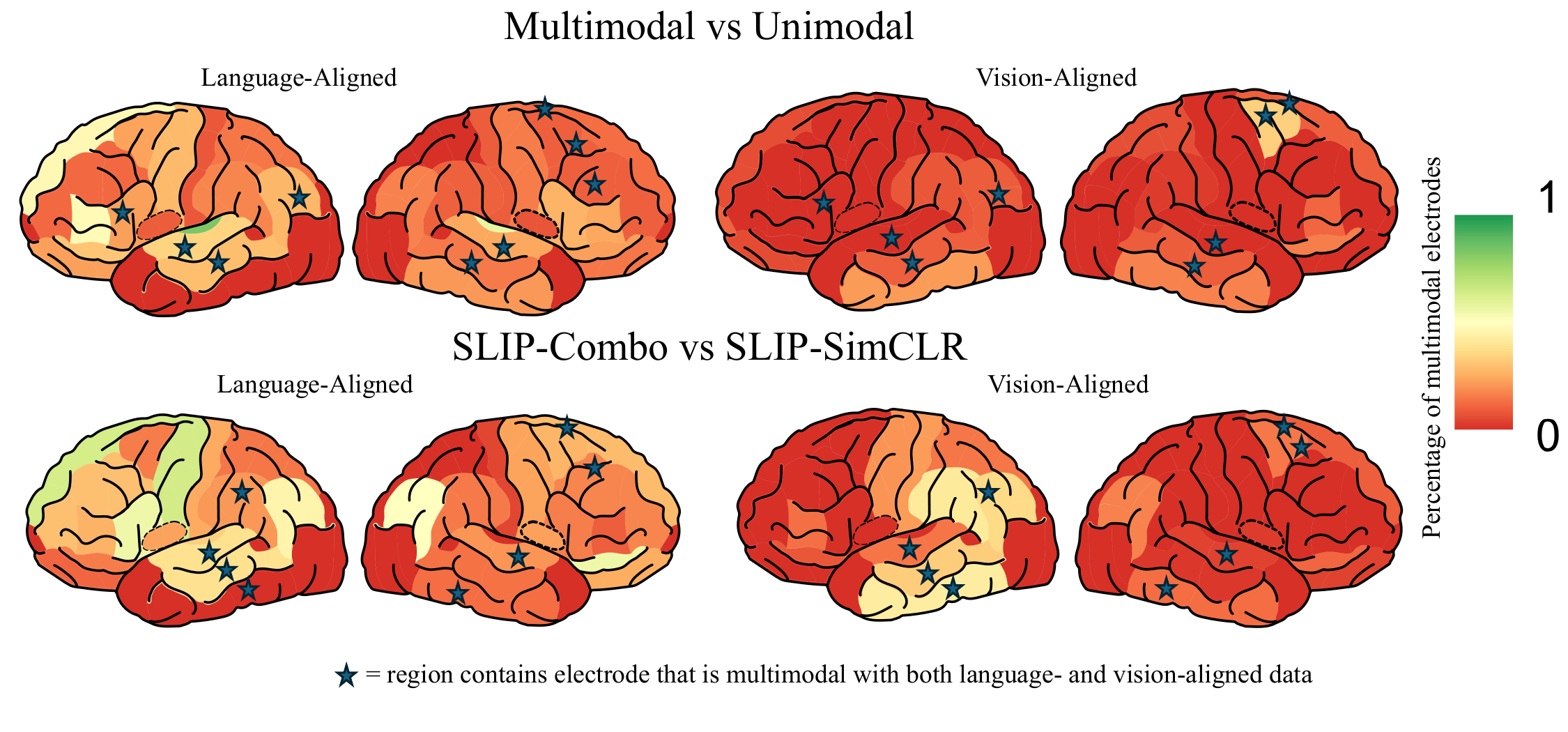}
    \caption{\textbf{Multimodal Integration by Region}. Here, we show candidate sites of multimodal integration aggregated into regions from the DKT atlas. For each site we compute the percentage of multimodal electrodes using the first test and the (left) language or (right) vision alignment. The top panel designates a site as multimodal if the best model that explains that electrode significantly outperforms all unimodal models. The bottom panel controls for architecture, parameters, and datasets by comparing SLIP-Combo and SLIP-SimCLR. Red regions have no multimodal electrodes. Regions which have at least one electrode that is multimodal both with the vision and language aligned stimuli are marked with a blue star. We notice that many electrodes occur in the temporoparietal junction with a cluster in the superior temporal cortex, middle temporal cortex, inferior parietal lobe, etc. Other areas we identify include the insula, supramarginal cortex, the superior frontal cortex, and the caudal middle frontal cortex.}
    \label{fig:mul_uni}
    \vspace{-1ex}
\end{figure*}

Using our multimodality tests to evaluate the predictive power of multimodal models against unimodal models across the two dataset alignments, we obtain the results shown in Table~\ref{tab:mul-tests}: The weak test of multimodality showed that 213/1090 (19.5\%) and 60/1090 (5.50\%) electrodes were more predictive using language- and vision-aligned event structures respectively, with average performance differences of $r_{\text{Pearson}} =0.082 [0.011, 0.21]$ and $0.081 [0.016, 0.344]$. The weak SLIP test yielded 218/1090 (20\%) and 73/1090 (6.70\%) electrodes for language- and vision-aligned structures, respectively, with performance differences of $r_{\text{Pearson}} = 0.046 [0.01, 0.140]$ and $0.024 [0.01, 0.04]$ between SLIP-SimCLR and SLIP-Combo vision transformers.

The strict test of multimodality found 12/1090 (1.1\%) electrodes were more predictive in both alignments, with average differences of $r_{\text{Pearson}} = 0.0766 [0.013, 0.163]$ and $0.0922 [0.019, 0.304]$. The strict SLIP test showed 28/1090 (2.57\%) electrodes favored the SLIP-Combo over the SLIP-SimCLR in both alignments, with differences of $r_{\text{Pearson}} =0.0522 [0.011, 0.10]$ and $0.026 [0.0162, 0.044]$. The non-linear integration test reiterated the 12/1090 electrodes from the third test, showing a consistent preference for multimodal models over MultiConcat and MultiLin, with performance differences of $r_{\text{Pearson}} = 0.0566 [0.025, 0.113]$ and $0.084 [0.029, 0.21]$ in the language- and vision-aligned datasets, respectively.

In examining the DKT atlas in Figure~\ref{fig:mul_uni}, it is evident that the largest cluster of multimodal electrodes is around the temporoparietal junction -- a result that aligns well with previous studies. Key regions include the superior and middle temporal cortex, the inferior parietal lobe, and the supramarginal gyrus, which are close and theoretically linked to vision-language integration. These areas, which prior work has found to be crucial for tasks like auditory-visual processing \citep{petrides2023evolution, bzdok2016left}, emotion processing, and social cognition \citep{stoeckel2009supramarginal}, support our findings and previous theories. The multimodal abstractions at this junction might explain their better prediction by multimodal representations. Additionally, electrodes passing tests 3 and 4 in the frontal and prefrontal cortex, specifically in the superior frontal lobe \citep{schilling2013cortical}, caudal middle frontal cortex, and pars orbitalis \citep{belyk2017convergence}, suggest complex cognitive processing in vision-language integration. This indicates a widespread brain network that may be involved in this integration, corroborating our results and existing literature which focused on more specific forms of vision-language integration.

Our multimodality tests demonstrate that multimodal models can greatly out-perform unimodal models at predicting activity in the brain, sometimes by close to $r_{\text{Pearson}} = 0.1$ at some electrodes. This underscores that multimodality could be an important factor in improving connections between deep networks and the brain. Furthermore, the areas we identify have commonly been associated with specific forms of vision-language integration identified in prior analyses. These prior analyses were constrained by smaller datasets with strong controls. We reduce these controls and still meaningfully identify the same areas for future study. More specifically, our analyses allow us to study vision-language integration without committing to a specific structural hypothesis. Despite this more general search space, we find meaningful overlap with prior work.

\subsection{Model Task Performance}
\begin{table}
\begin{tabular}{lc}
\toprule
\textbf{Model}    & \textbf{Next-Word Perplexity ($\downarrow$)}        \\ \midrule
Average Unimodal   & 133.4                             \\
Average Multimodal & 210.3                             \\ \midrule
                   & \textbf{Scene Class Accuracy ($\uparrow$)} \\ \midrule
Average Unimodal   & 74.2                              \\
Average Multimodal & 54.3                              \\ \bottomrule
\end{tabular}
\caption{\textbf{Multimodal vs Unimodal Task Performance.} We report average unimodal task performance for unimodal models and multimodal models. We show next-word prediction perplexity and scene-cut class accuracy for one movie. Our findings demonstrate that unimodal models have better unimodal representations than multimodal models as reflected by better performance.}
\vspace{-1ex}
\label{tab:task-perf}
\end{table}

While our study aims to explore vision-language integration, we must consider other explanations, such as whether our multimodal networks outperform unimodal networks in language or visual reasoning. This could imply that our findings are due to incremental improvements in unimodal processing, rather than meaningful vision-language integration. To address this, we evaluated our multimodal networks' performance on unimodal tasks. For language tasks, we assessed next-word prediction performance on our movie dialogue in both multimodal and unimodal language networks, using perplexity as a metric. For vision tasks, we tested scene classification abilities using Places365. 

Our results are detailed in Table~\ref{tab:task-perf}. We show that multimodal networks perform worse on unimodal tasks compared to unimodal networks. This reduces the likelihood that our findings are merely associated with improvements in unimodal representations or unimodal processing by multimodal networks, and provides more evidence that our discoveries are associated with some form of vision-language integration.

\subsection{Which multimodal model is most `brain-like'?}
\begin{figure*}[t]
    \centering
    \begin{tikzpicture}
        \node at (0.25, 4.8) {\textcolor{black}{\LARGE Language-Aligned}};
        \node[inner sep=0pt] (image) at (0,0) {\includegraphics[width=0.7\textwidth]{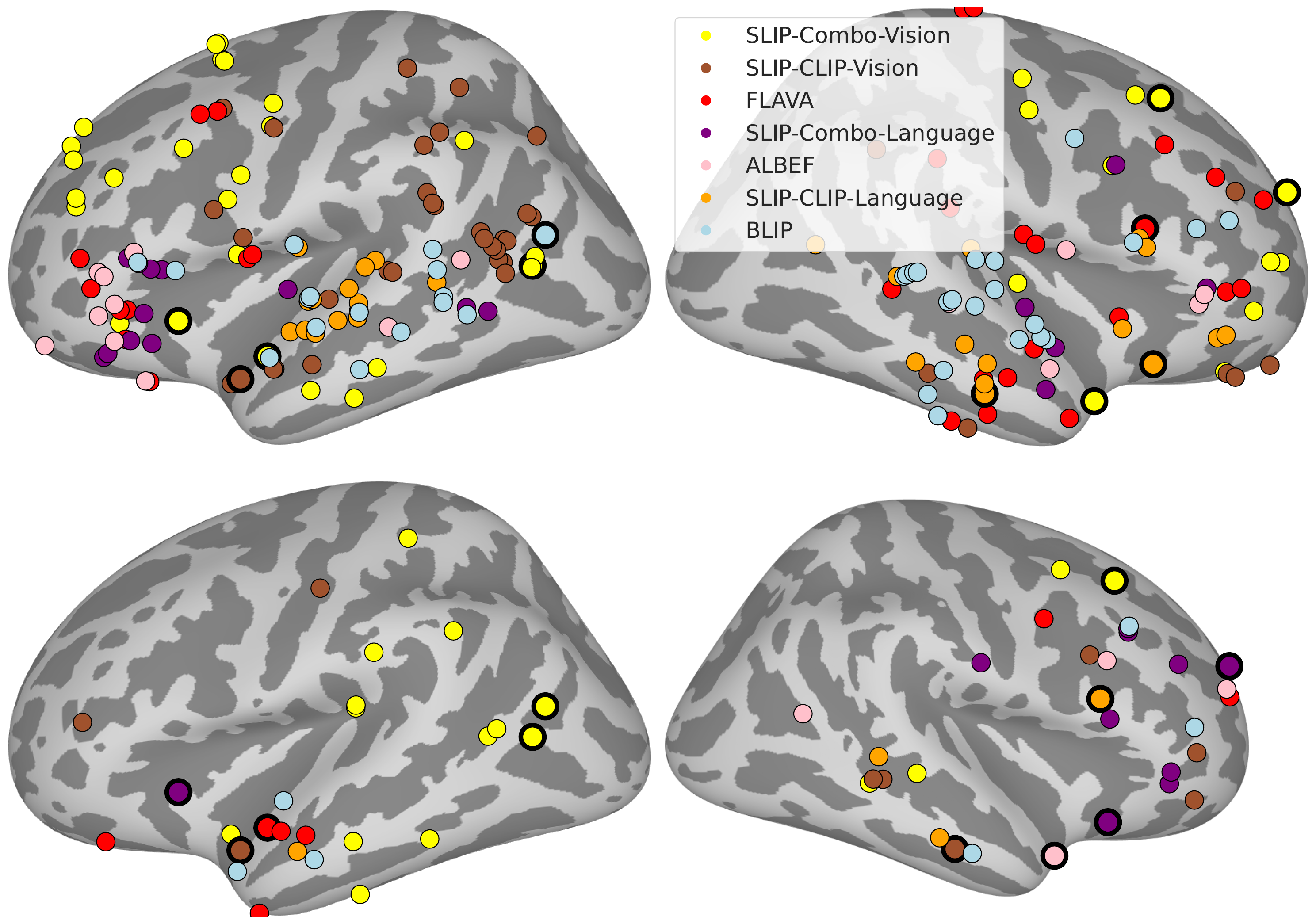}};
        \node at (0.25, 0.03) {\textcolor{black}{\LARGE Vision-Aligned}};
    \end{tikzpicture}
    \caption{\textbf{Best Models of Multimodal Integration}. Here, we visualize the individual electrodes that pass our weak and strict multimodality tests for the language-aligned (top, 213 electrodes) and vision-aligned datasets (bottom, 90 electrodes), adding a bold outline to electrodes that pass across both datasets (12 electrodes). We color the electrodes by the top-ranked multimodal model that predicts activity in the electrode. We see that models such as SLIP-Combo and SLIP-CLIP often predict activity the best across datasets. We also see that BLIP and Flava are the best architecturally multimodal models.}
    \label{fig:mul-arch}
\end{figure*}

In Figure~\ref{fig:mul-arch}, we use our model ranking techniques and show the most predictive multimodal model for each of the electrodes that pass our tests of multimodality (specifically, the weak test of multimodality, the strict test of multimodality, and the non-linear integration test). We see consistently that trained multimodal models such as SLIP-Combo and SLIP-CLIP predict the brain better than architecturally multimodal models such as ALBEF, BLIP, or Flava.

We also find that the SLIP-Combo and SLIP-CLIP language encoders are the most predictive networks in the vision-aligned data. This may indicate that there exists a notion of ``concepts'' that describe stimuli at a higher level than simply the vision or text modality, and that the SLIP-* models align well with the brain's representations of such concepts.

There are many possible reasons why models like SLIP-Combo or SLIP-CLIP out-predict architecturally multimodal models; these include dataset considerations or the need for better cross-attention design. However, architecturally multimodal models do seem in general to do a better job of predicting the language-aligned activity. Among the architecturally multimodal models, BLIP is the most predictive of the brain. These results could indicate that the cross-attention mechanism used in models such as BLIP is better at integrating vision and language in a manner similar to the brain. There is room for future work in this direction, where we focus on specific motifs/network designs and how they associate with vision-language integration in the brain.

In general, our findings show that network parameter size does not correlate with predictivity. SLIP-Combo and SLIP-CLIP have fewer parameters than our architecturally multimodal models and even our unimodal models. This indicates a special feature in CLIP-style training that could be studied more carefully in future work.

\section{Conclusion}
The methodology introduced here provides a fine-grained analysis that overcomes a first hurdle: it distinguishes randomly initialized and trained language networks in every modality individually and then across modalities. Having overcome this hurdle, we can now identify areas which are better explained by multimodal networks compared to unimodal networks and linearly-integrated language-vision networks. The most-fine grained result we provide compares SLIP-Combo vs SLIP-SimCLR, a multimodal and unimodal network controlling for architecture, dataset, and parameter count. We release a toolbox for multimodal data analysis along with, upon request, the raw neural recordings under a permissive open license such as Creative Commons.

We identified a cluster of sites which connect vision and language. This appears to be a network which spans the temporoparietal junction, connecting the superior temporal cortex, middle temporal cortex, inferior parietal lobe, and supramarginal gyrus, to areas in the frontal lobe, containing the pars orbitalis, superior frontal cortex, and the caudal middle frontal cortex. These areas align with prior studies and analyses on particular aspects of vision-language integration in the brain.

While our data has high fine-grained temporal resolution, and our method is sensitive to the time course of the signal, our final analysis aggregates across time bins. We have not investigated how multimodal integration occurs as a function of time. This could be used to derive a time course of integration across the brain and to establish a more specific network structure.

Our method is agnostic to the modalities used, or to the number of modalities. Neural networks exist which integrate not just language and vision, but also audio \citep{perez2020audio, gong2022contrastive} and motor control \citep{hu2023pre, li2023learning}. These could also be used with our method (our data even explicitly enables future modeling with audio). The distinction and interaction between audio processing and language processing could be revelatory about the structure of regions which transition from one to the other, like the superior temporal gyrus.

We plan to investigate which network computations, motifs, and hyper-parameters are most brain-like when designing multimodal networks. In particular, we will aim to identify the types of modules (e.g., cross-attention with language attending vision or vision attending language) that best fit the brain and the principles of designing multimodal datasets. Our approach could be used to determine which of these approaches is most brain-like, helping to guide future research. One could even imagine this approach enabling a variant of Brain-Score dedicated to multimodal processing in order to systematically investigate this over time as a community.

\textbf{Limitations}. What these areas of integration do is unclear. A causal and mechanistic understanding that relates areas to one another will be required to address that question --- one that our single-electrode analysis cannot provide. Extending the tools we built for that case will not be simple and it is unclear whether we have enough data at present to run this type of analysis. One weakness of our approach is that it relies on models, and models can change dramatically over time. It may be that the areas of integration our analysis finds will change as models improve. As with any model-based method this can make interpretation difficult.

\section*{Impact Statement}
Issues of sensory integration are involved in disorders like autism spectrum disorder \citep{doumas2016postural}. Developing a model-based mechanistic understanding of how the neurotypical brain builds multimodal representations may eventually allow us to develop better models of neurodivergence. At the same time, better understanding how the brain in general integrates multimodal information may be a key step to solving many lingering issues that exist with multimodal reasoning and inference in current state-of-the-art AI/ML models. For example, many multimodal algorithms continue to suffer from issues of relational and compositional reasoning \citep{zhang2024m3exam, conwell2022testing} that may be linked to the absence of more brain-like (or at least brain-comparable) multimodal integration. 

\textbf{Compute and Data Statement} This work was carried out under the supervision of MIT's Institutional Review Board (IRB). 28 modern GPUs on 7 machines were used for four weeks, evenly distributed across experiments.

\section*{Acknowledgements}
This work was supported by the Center for Brains, Minds, and Machines, NSF STC award CCF-1231216, the NSF awards 2124052 and 2123818, the MIT CSAIL Machine Learning Applications Initiative, the MIT-IBM Watson AI Lab, the CBMM-Siemens Graduate Fellowship, the DARPA Artificial Social Intelligence for Successful Teams (ASIST) program, the DARPA Knowledge Management at Scale and Speed (KMASS) program, the DARPA Machine Common Sense (MCS) program, the United States Air Force Research Laboratory and the Department of the Air Force Artificial Intelligence Accelerator under Cooperative Agreement Number FA8750-19-2-1000, the Air Force Office of Scientific Research (AFOSR) under award number FA9550-21-1-0014, NIH grant R01EY026025, and the Office of Naval Research under award number N00014-20-1-2589 and award number N00014-20-1-2643. The views and conclusions contained in this document are those of the authors and should not be interpreted as representing the official policies, either expressed or implied, of the Department of the Air Force or the U.S. Government. The U.S. Government is authorized to reproduce and distribute reprints for Government purposes notwithstanding any copyright notation herein.

\bibliography{citation/deepneuro, citation/current}
\bibliographystyle{icml2024}

\newpage
\appendix
\onecolumn
\section{Neural Data Details}
We provide details of our neural data recordings and preprocessing steps taken to form \emph{event structures}. 
\label{appendix:neural-event-structures}

\subsection{Data Collection Overview}
\begin{figure}[h]
    \centering
    \includegraphics[width=\textwidth]{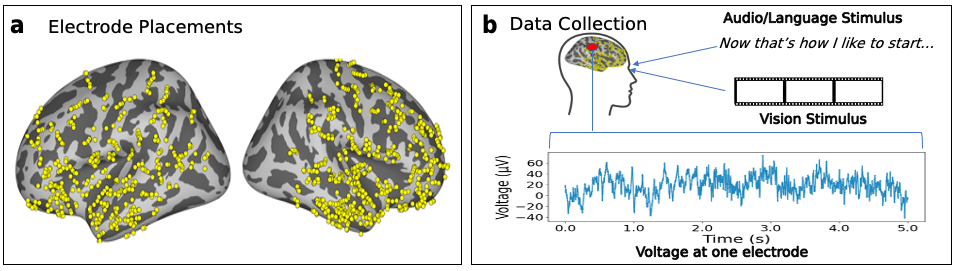}
    \caption{\textbf{Data Overview}. (a) The electrode placements over all subjects. Each yellow dot denotes an electrode collecting invasive field potential recordings for further analysis in our experiments. (b) An overview of our data collection procedure. Subjects are presented feature length films while neural data is collected from these electrodes in the brain. }
    \label{fig:data-overview}
\end{figure}

We provide an overview of our dataset in Figure~\ref{fig:data-overview}. We provide a subject-by-subject breakdown in Table~\ref{tab:subject-stats}, providing information on the age, number of electrodes, and the length of the recording.

\begin{table}[h]
\centering
\begin{tabular}{ccp{1cm}p{4.5cm}p{1.5cm}p{1cm}}
\toprule
\textbf{Subj}. & \textbf{Age (yrs.)} & \textbf{\# Electrodes} & \textbf{Movie} & \bf Recording time (hrs)\\
\midrule
1                 & 19 & 154 & Fantastic Mr. Fox                               & 1.83 \\             
\midrule
2                 & 12 & 162 & Venom                               & 2.42 \\  
\midrule
3                 & 18 & 134 & Cars 2                           & 1.92 \\ 
\midrule
4                 & 6 & 156 & Fantastic Mr. Fox                      & 1.5  &   \\
\midrule
5                & 16  & 162  & Sesame Street Episode                  & 1.28 &   \\
\midrule
6                & 4.5  & 106 & Ant Man                                & 2.28 &   \\
\midrule
7          & 12       & 216 & Cars 2                                 & 1.58\\
\bottomrule
\end{tabular}
\caption{\textbf{Subject statistics}
Age (second columns), number of electrodes (third column), movie shown (fourth column) and recording time (fifth column) per subject. Electrode placements are done for clinical purposes and the distribution of electrode locations differ from subject to subject. 
The average amount of recording data per subject is 1.83 (hrs).
}
\label{tab:subject-stats}
\end{table}

\subsection{Event Structures}
\begin{figure}
    \centering
    \includegraphics[width=0.9\textwidth]{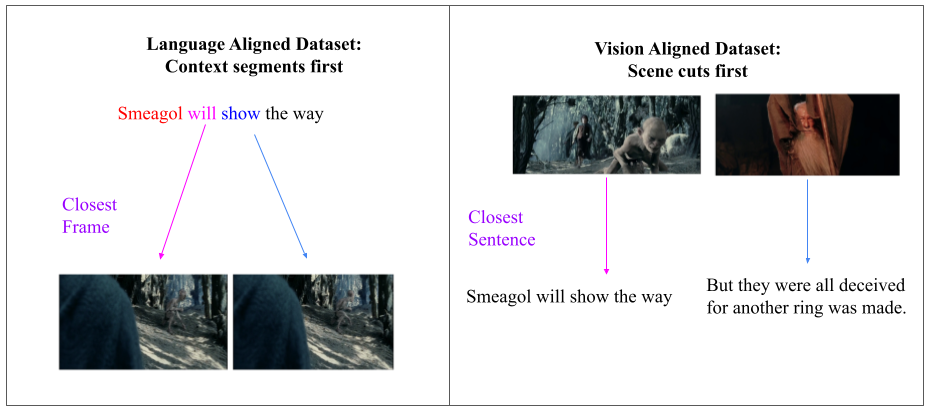}
    \caption{\textbf{Event Structures}. We construct two datasets for understanding vision-language integration in the brain. (Left) The language-aligned dataset consists of choosing context segments and the closest frame. Context segments choose each word and the corresponding sentence context. (Right) The vision-aligned dataset consists of choosing a scene cut and the closest sentence that occurs after a scene cut.}
    \label{fig:event-structures}
\end{figure}

We parse our neural activity into individual language and single movie-frame combinations (which we call interchangeably \emph{event structures} or \emph{text-image pairs}) by discretizing the movie stimulus, allowing us to feed inputs to our deep neural network models (which are not trained on movie data). We define event structures by the guiding feature used to select a particular text-image pair in the movie for analysis. So as not to unfairly prioritize one modality over the other or impose a hypothesis over vision-language integration, we design two different kinds of event structures: The first kind of event structure consists of word onset times, a language-aligned event. Word onsets have been used in prior work \citep{goldstein2021thinking} and are commonly associated with language processing. For each word onset, we take the prior sentence context of the given word to add contextual information for the language models. We also take the closest frame after the word onset as the associated image input. The second kind of event structure consists of visual scene cuts (i.e. camera cuts). We extract the frames associated with a scene cut as a proxy for visual processing given a shift in the pixel distribution between frames. We then take the closest sentence that occurred after the scene cut. (Note that by language-alignment or vision-alignment, here, we mean the anchoring of points in neural time-series to points in the movie).

We use these two kinds of event structures to create two datasets. Our language-aligned dataset consists of [context of a given word, closest frame pairs] with the associated neural activity as processed in Section~\ref{sec:methods}. Our vision-aligned dataset consists of [scene cut frames, closest sentence to a scene cut frame] with similar processing on the neural activity. We analyze all results over the datasets individually and then compare results across the datasets to identify electrodes for multimodal integration.

We note that our two datasets cover many possible hypotheses of vision-language integration. The language-aligned dataset likely covers short-term integration as each corresponding context segment has a nearby frame. However, our vision-aligned dataset likely covers long-term integration since there is separation between the scene cut and corresponding sentence. This design makes our experiments more difficult by considering many forms of possible visio-linguistic reasoning.

\subsection{Notes on Stimulus Independence (Autoregression)}

Converting neural activity measured in response to naturalistic movie-viewing to a dataset of nomimally IID event-structures presents a particular challenge often explicitly avoided in experimental designs that leverage otherwise unrelated natural images or language prompts: that is, nonindependence in the form of autoregression. Movies (driven as they are by common visuolinguistic themes) contain inherently autoregressive structure that can lead to overfitting in parametrized predictive models designed to predict neural response patterns evoked by that structure. The parsing of our final event-structures into training, testing and validation splits was designed explicitly to assess for such overfitting. When creating the train-validation-test splits, we assign contiguous chunks of the movie to each split. In practice, and especially for movies with more linear narrative structure, we assumed this contiguous splitting could provide at least a weak form of independence between sampled event-structures. While this by no means fully accounts for the non-independence of the stimulus set writ large, our results across the training, validation, and test splits suggest that it does help to minimize potential overfitting. In future work, we hope to revisit our event-structure delineation and sampling, potentially leveraging movie-trained models like Salesforce's ALPRO \citep{li2022align} to select stimuli that are more distinct not just at the level of pixels or words, but in latent feature space.

\section{Candidate Deep Neural Network Models}
\begin{table}[h]
\resizebox{\textwidth}{!}{%
\begin{tabular}{lcccc}
\toprule
\bf Model               & \bf Modality                   & \bf Architecture  & \bf Parameters & \bf Dataset \\ \midrule
ALBEF               & Architecturally Multimodal & Transformer   & 209.8M & Conceptual Captions/SBU Captions/COCO/Visual Genome/Conceptual-12M     \\
BLIP                & Architecturally Multimodal & Transformer   & 223.5M &  Conceptual Captions/SBU Captions/COCO/Visual Genome/Conceptual-12M     \\
Flava               & Architecturally Multimodal & Transformer   & 241.4M & Public Multimodal Datasets      \\ \midrule
SLIP-Combo Vision   & Trained Multimodal         & Transformer   & 86M &   YFCC15M      \\
SLIP-Combo Language & Trained Multimodal         & Transformer   & 63.6M & YFCC15M      \\
SLIP-CLIP Vision    & Trained Multimodal         & Transformer   & 86M &   YFCC15M      \\
SLIP-CLIP Language  & Trained Multimodal         & Transformer   & 63.6M &  YFCC15M     \\ \midrule
SBERT               & Language                   & Transformer   & 109.5M & SNLI/Multi-Genre NLI     \\
SimCSE              & Language                   & Transformer   & 109.5M & $\text{QQP}^4$/Flickr30K/ParaNMT/SNLI     \\ \midrule
SLIP-SimCLR         & Vision                     & Transformer   & 86M &  YFCC15M      \\
BEIT                & Vision                     & Transformer   & 86.3M &   ImageNet-1K    \\
ConvNeXt            & Vision                     & Convolutional & 109.9M &  ImageNet-1K   \\ \bottomrule
\end{tabular}%
}
\caption{\textbf{Multimodal and Unimodal DNNs}. A catalogue of the networks we include in this experiment. We compare architecturally multimodal networks, trained multimodal networks, unimodal language and unimodal vision networks. We mostly study transformers but also include ConvNeXt, a CNN model. We tabulate the number of parameters in the model.}
\label{tab:networks}
\end{table}
\label{appendix:networks}
We present a full set of networks in Table~\ref{tab:networks}. Because they control for dataset and architecture (varying only the learning objective), comparisons amongst the variants of the SLIP models are our most empirically rigorous test of multimodality.

However, given that the SLIP models contain only one kind of multimodal - unimodal contrast (SLIP-SimCLR versus SLIP-Combo's visual encoder), we added a number of uncontrolled model contrasts to assess the predictive power of unimodal and multimodal representations more generally. These models include ALBEF \citep{li2021align} (a two channel multimodal encoder that uses a vision transformer and language transformer trained with a contrastive loss followed by a multimodal transformer); BLIP \citep{li2022blip} (a two channel multimodal enoder similar to ALBEF but trained with an image-text matching loss and momentum model); Flava \citep{singh2022flava} (a two channel multimodal encoder with a multimodal encoder that builds fused embeddings and trained reconstructively); SBERT \citep{reimers-2019-sbert} (a unimodal masked language transformer for sentence embeddings); BEIT \citep{bao2021beit} (a unimodal vision transformer trained via masked image reconstruction); SimCSE \citep{gao2021simcse} (a unimodal language transformer trained via contrastive learning); ConvNeXt \citep{liu2022convnet} (a unimodal vision convolution network built by modifying the ResNet architecture). These models provide a broader sample of multimodal and unimodal networks, while still maintaining some core similarities with the SLIP models (transformer backbones or contrastive learning.)

In Table~\ref{tab:networks}, we also provide the training data used when pre-training these networks. In general, we find that all models are trained on open datasets which contain more naturalistic dataset. For example, most vision models such as BEIT or ConvNeXt train on ImageNet \citep{deng2009imagenet} which contains real-world images. Similarly, our multimodal models train on YCCFM-15M or Conceptual Captions \citep{sharma2018conceptual} which contains images-text pairs of media objects that are more realistic. This might impact performance of our networks in building representations of movie frames and dialogue. However, our language networks, SBERT and SimCSE, train on SNLI \citep{bowman2015large} and Multi-Genre NLI \citep{williams2018broad}, datasets that contain fictional references. This further supports our design. Our unimodal networks have potentially better performance than our multimodal networks when comparing performance to the brain.

We also introduce two \emph{linearly integrated vision-language models}, \emph{MultiConcat}, and \emph{MultiLin}. MultiConcat consists of concatenating representations from SimCSE and SLIP-SimCLR, and MultiLin extends MultiConcat by introducing a trained linear projection to project the concatenate representation to a dense vision-language vector trained using the NLVR-2 dataset \citep{suhr2018corpus}. By introducing these models, we aim to distinguish between areas that are simply responding to the presence of vision and language features and areas that are integrating vision and language in a rich, non-linear fashion using comparisons we describe more in detail.

We assess both trained and randomly-initialized versions of these models first and foremost because, in most cases, the multimodality of these models is a function ONLY of their learning objective: This means, for example, that models like the SLIP models -- which consist of architecturally encapsulated vision and language encoders -- cannot, in the absence of training, be considered multimodal. Models like ALBEF, BLIP, or Flava, on the other hand, may be considered multimodal even in the absence of training due to architectural inductive biases such as cross-modal attention-heads that integrate linguistic and visual inputs from the outset of processing. 

\section{Neural Regressions}
\begin{figure*}
\centering
    \includegraphics[width = 0.7\textwidth]{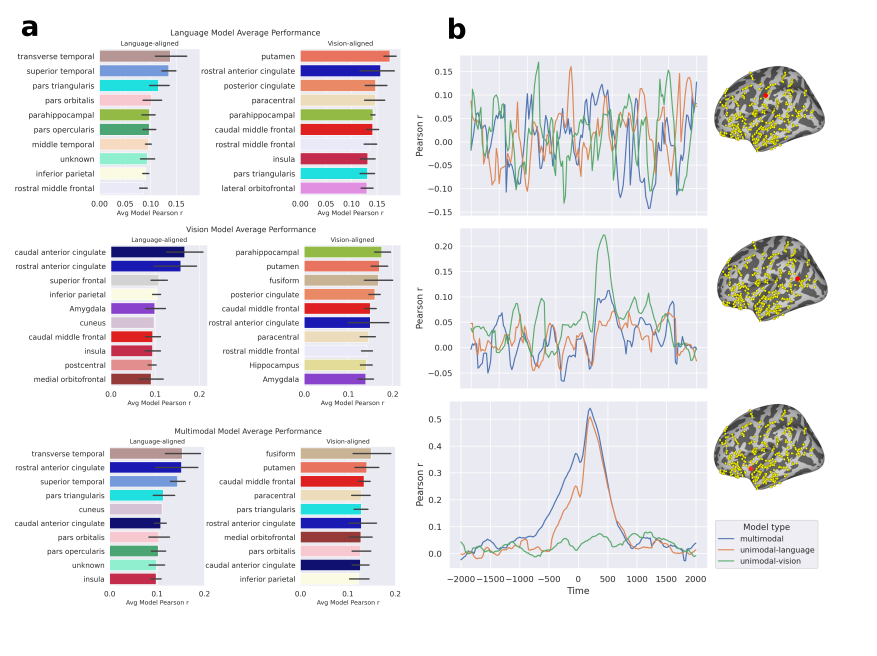}
    \vspace*{-5ex}
    \caption{\textbf{Results Overview}. (a) The top ten Desikan-Killiany-Tourville (DKT) regions, ranked based on average predictivity, Pearson r, across electrodes in that region. Error bars represent the standard error associated with averaging the Pearson r over the electrodes and time bins. All results shown here are from pretrained variants of the model. (top) A language model, SBERT. (mid) A vision model, SLIP-SimCLR. (bottom) A multimodal model, BLIP. (b) The predictivity of these same three models for three typical electrodes. (top) An arbitrary electrode which is not responsive to language or vision. (mid) An electrode which is responsive to vision, but not language, and is not better explained by multimodal integration. (bottom) An electrode which is responsive to language and that is better explained by a multimodal network. Confidence intervals are not shown for clarity.}
    \label{fig:new_results}
    \vspace*{-2.5ex}
\end{figure*}
\label{appendix:regression}

In this section, we detail our neural regression pipeline, which proceeds in 4 phases: feature extraction, dimensionality reduction (via sparse random projection), cross-validated ridge regression, and scoring.

\subsection{Feature Extraction}
\label{sec:fe}
This follows from approaches taken in \citet{conwell2021neural}. We consider feature extraction to mean the extraction of a separate feature vector at \textit{every layer} in a network -- in other words, each distinct tensor operation module that progressively transforms model inputs into outputs. This means, for example, that we consider not only the outputs of each transformer attention head, but also of the individual key, query, value computations that produce them. If the layer has more than 1 dimension, then we flatten the tensor such that each layer represents any given input as a 1-dimensional feature vector. (Note: This flattening makes no assumptions about the separation of a given feature space into spatial and semantic components, and allows the subsequent regression to reweight all contributing components as relevant). The output tensor thus constitutes a dataset of $n$ inputs (either images, sentences, or image-sentence pairs) as an array $\mF \in \mathbb{R}^{n\times D}$ where $D$ is the dimensions of the feature vector.

\subsection{Sparse Random Projection}
\label{sec:srp}
For certain flattened feature vectors, the dimensionality $D$ is very large, and as such performing ridge regression on $\mF$ is prohibitively expensive, with at best linear complexity with $D$, specifically $\mathcal{O}(n^2D)$ \citep{hastie2004efficient}. We use the Johnson-Lindenstrauss lemma \citep{johnson1984extensions, dasgupta2003elementary} to project $\mF$ to a low dimensional representation $P\in\mathbb{R}^{n \times p}$ that preserves pairwise distances in $\mF$ with errors bounded by a factor $\epsilon$. If $u$ and $v$ are any two feature vectors from $\mF$, and $u_p$ and $v_p$ are the low-dimensional projected vectors, then
\begin{equation}
    \label{eq:johnson}
    (1-\epsilon)||u - v||^2 < ||u_p - v_p||^2 < (1+\epsilon)||u - v||^2
\end{equation}
Equation~\ref{eq:johnson} holds provided that $p \geq \frac{4\ln(n)}{\epsilon^2/2 - \epsilon^3/3}$ \citep{achlioptas2001database}. To find the mapping from $\mF$ to $\mP$, we used \textit{sparse random projections} (SRPs) following \citet{li2006very}. The authors show a $\mP$ satisfying Equation~\ref{eq:johnson} can be found by $\mP = \mF\mR$ where $\mR$ is a sparse $n \times P$ matrix with i.i.d. elements shown below:
\begin{equation}
    r_{ij} = 
    \begin{cases}
        \sqrt{\frac{\sqrt{D}}{p}} & \text{with prob. } \frac{1}{2\sqrt{D}} \\
        0 & \text{with prob. } 1-\frac{1}{\sqrt{D}} \\
        -\sqrt{\frac{\sqrt{D}}{p}} & \text{with prob. } \frac{1}{2\sqrt{D}}
    \end{cases}
\end{equation}

If $\mF$ has dimensionality $D$ that is less than the dimensionality of the Johnson-Lindenstrauss lemma, then no projection is applied. In this case, $\mP = \mF$. 

\subsection{$k$-fold Ridge Regression}
To determine how well vision and language networks predict activity in the brain, we ran regressions from representations extracted from a specific layer of either a multimodal or unimodal network to predict the average activity of the SEEG signals over a window of time for all electrodes of our $7$ subjects. We detail the steps we took to run regressions per subject below.

We use ridge regression to predict the average activity, $\vy$, at a given electrode and time point as constructed in Section~\ref{sec:methods}, from their associated DNN features $\mP$. Given the sequential nature of our data, we used a $5$-fold cross-validation procedure. For each fold, we split our dataset of representations into a contiguous training set($80\%$), $\mP_{\text{train}}$ and $\vy_{\text{train}}$, a contiguous validation set ($10\%$), $\mP_{\text{valid}}$ and $\vy_{\text{valid}}$, and contiguous testing set ($10\%$), $\mP_{\text{test}}$ and $\vy_{\text{test}}$. Each split takes a contiguous chunk of event structures in order of their occurrence in the movie, and each fold changes the starting point of the training, validation, and testing set such that different contiguous chunks are assigned to a different set. We standardize the columns of $\mP_{\text{train}}$ and $\mP_{\text{valid}}$ to have mean $0$ and a standard deviation of $1$ and fit this standardization on $\mP_{\text{test}}$. We fit the coefficients $\hat{\beta_i}$ of a regression model on the train dataset such that $\vy_{\text{train}} = \mP_{\text{train}}\hat{\beta}_i + \epsilon$ with minimal error $||\epsilon||$. Ridge regression penalizes large $||\hat{\beta}||$ proportional to a hyperparameter $\lambda$, which is useful in preventing overfitting when regressors are high-dimensional and highly correlated. Each $\hat{\beta}$ is calculated by the fixed ridge regression solution:

\begin{equation}
    \label{eq:ridge}
    \hat{\beta} = ((\mP_{\text{train}})^T\mP_{\text{train}} + \lambda\mI_d)^{-1}(\mP_{\text{train}})^T\vy_{\text{train}}
\end{equation}

The coefficients $\hat{\beta}$ are then used to predict the held out data where:
\begin{equation}
\begin{gathered}
    \hat{\vy_{\text{valid}}} = \mP_{\text{valid}}\beta \\
    \hat{\vy_{\text{test}}} = \mP_{\text{test}}\beta
\end{gathered}
\end{equation}

We use the \textit{KFold} function from \citet{pedregosa2011scikit} and implemented ridge regression in Pytorch \citep{paszke2019pytorch}. In this analysis we run the $5$-fold regression per $\lambda$ value, where $\lambda$ was varied using a logarithmic grid search over $10^{-1}$ to $10^{6}$. On each fold, we calculated a \textit{score} for the prediction $\hat{\vy_{\text{valid}}}$ and $\hat{\vy_{\text{test}}}$ by computing the Pearson correlation coefficient. This score is averaged over the $5$ folds to get final validation and test set scores. We choose the best $\lambda$ value using the cross-validated scores and take the associated test scores with the $\lambda$ value. We run this regression for all electrodes and time points simultaneously.

To analyze network performance over all layers, we select the best performing layer using the validation set. Specifically, per electrode, we average the validation correlation scores over time and take the layer with the max average score. We then take the associated test set correlation score as the overall score per model. 

We provide initial results in Figure~\ref{fig:new_results}. We first show the top ten Desikan-Killiany-Tourville (DKT) regions based on the average score over one model from the three model types: a unimodal language model, a unimodal vision model, and a multimodal model. We notice several re-occurring regions across the three model types including the superior temporal cortex, the rostral anterior cingulate, the transverse temporal cortex, the fusiform and the superior frontal cortex. These are regions associated with the temporoparietal junction, indicating a potential network of vision-and-language processing with connections to the frontal lobe. Many of these regions are associated with high level emotion processing and spatial processing. Furthermore, we show the results of the three model types on three particular electrodes. The first electrode is in the precentral gyrus and gives the average performance of the models with peak performance of $r_{\text{Pearson}} = 0.15$. The second electrode is in the inferior parietal lobe and gives an example of an electrode with strong performance from vision models. Finally, the last electrode is in the superior temporal lobe and gives an example of an electrode processing language and potentially integrating multimodal features. The analysis will use these scores over time per electrode and compare the score distributions to quantify statistically whether a multimodal model is performing significantly better than a unimodal model.

\section{Statistical Testing Details}
\label{appendix:multi-test}

\subsection{Bootstrapping over Event Structures}
\begin{figure}
    \centering
    \includegraphics[width=\textwidth]{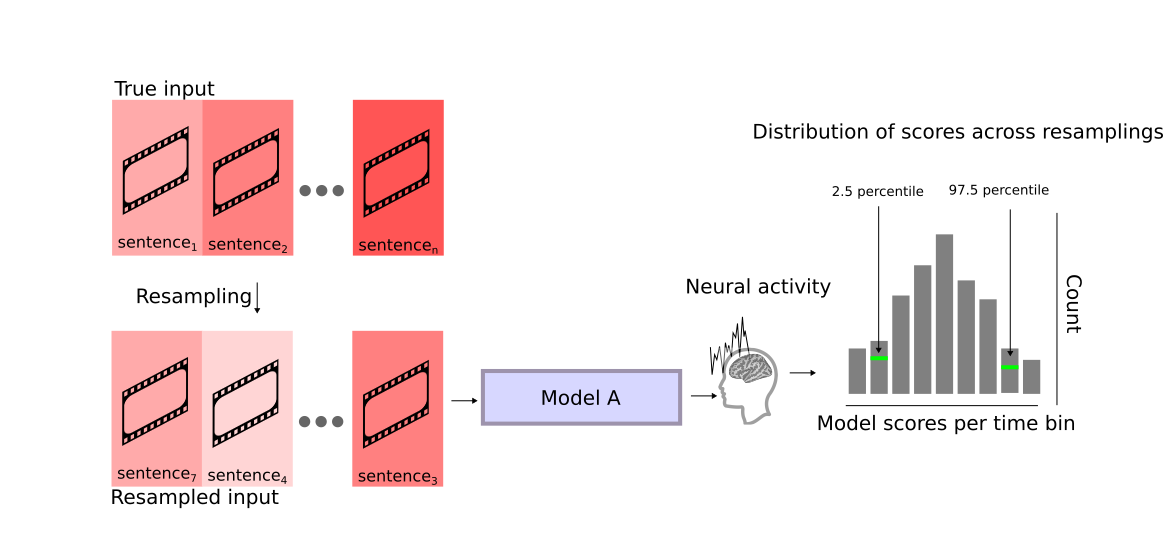}
    \caption{\textbf{Boostrapping Procedure}. We bootstrap our event structures and corresponding neural activity to derive 95\% confidence intervals per time bin per electrode on our training, validation, and testing set in our analysis.}
    \label{fig:bootstrapping}
\end{figure}
As shown in Figure~\ref{fig:bootstrapping}, we introduce a bootstrapping procedure over our event structures (image-text pairs) which allows us to derive confidence intervals on our regression per time bin $t$ and electrode $e$ in our analysis. 

In our bootstrapping procedure, we first resample the image-text pairs and corresponding neural activity with replacement per movie and subject 1000 times. We use the same 1000 resampled image-text pairs and corresponding neural activity across all models (both trained and randomly initialized) to allow for model comparison in later analysis.  We denote this resampling with replacement as follows:
\begin{equation}
    \{(\mP^{(1)}), \vy^{(1)}), \cdots (\mP^{(999)}), \vy^{(999)}), (\mP^{(1000)}), \vy^{(1000)})\} \sim \{\mP, \vy\}
\end{equation}

Most importantly, for each $(\mP^{(i)}, \vy^{(i)})$, we sort the resampled indices by their occurrence in the movie time to maintain the autoregressive structure of data. We note that the scores are skewed upwards when we do not sort.

We then rerun the regression for each $(\mP^{(i)}, \vy^{(i)})$ to derive the 95\% confidence intervals by taking the 2.5th percentile and 97.5th percentile on our score distribution per $t$ per $e$ on the training, validation, and testing set. Upon inspection of bootstrapping scores on a sample of time bins and electrodes, we note that the scores are normally distributed. 

We use this information as a filter, identifying time-bins and electrodes that do not have meaningful response to our event structures. We interpret the lower confidence interval of our validation set bootstrapping scores as the lower bound on our score parameter \footnote{We emphasize that this is done on our validation set to allow for unfiltered comparison on our test set scores.}. For a time bin where this lower bound is below 0, we can say that no meaningful mapping to the brain has been learned.

\subsection{Model Comparisons}
Each of our multimodality tests in the main analysis is predicated on a model comparison procedure. After our bootstrapping procedure, we filter the time bins in electrodes for models where the lower 95\% confidence interval of the validation score overlapped with 0 (see above). Furthermore, we remove models from further analysis on a particular electrode if all time bins of the model has a validation confidence interval that overlaps with $0$ for all time bins. 

Per electrode $e$, we identify models that have at least $10$ time bins where the lower validation confidence interval is greater than zero. If there are no other models with 10 such time bins, then only this model remains for further analysis and we refer to such models as "default winners", meaning the model has the default best performance on the electrode and has statistically significant performance on an electrode. 

For electrodes that did not have a "default winner", we design a model ranking based on a comparison of the mean bootstrapped validation score over time. We define a confidence interval on the bootstrapped validation scores over time per model on each electrode and sort the mean bootstrapped validation scores over time to obtain a model ranking. 

Using the first and second highest ranking models, we use a second-order bootstrapping procedure across time bins in $e$ rather than event structures. We only compare the top two models as these are the best predictors of the activity in $e$ and we assume that if the first ranking model is significantly better than then second ranking model, then this significant difference will hold with all other models. We calculate the difference in the average score across the 10 or more resampled time bins between the 2 candidate models in a given comparison. This procedure gives us a p-value per model comparison on each electrode. We repeat this procedure for all electrodes for each of the model comparison tests we describe below. We then use FDR (Benjamini-Hochberg) \citep{thissen2002quick} multiple comparisons corrections to adjust the $p$-value associated with each test on each electrode. 

Each of the 5 tests we conduct are suggestive of multimodality, but each successive test provides additional evidence. For tests 1 and 2, we only consider results per dataset alignment. Test 1 considers all comparisons but test 2 controls for architecture and training details with SLIP-Combo and SLIP-SimCLR. Since results in tests 1 and 2 could be explained by unimodal task performance, we introduce tests 3 and 4 where we identify sites that are multimodal in both dataset alignments. Such sites must be predictive by models that perform vision tasks and language tasks. In particular, we can note that test 3 and 4 are identifying sites where multimodal models outperform all unimodal language and unimodal vision models on all possible vision and language tasks associated with a particular neural site. We introduce test 5 as a test for finding the comparing types of multimodal integration, i.e. comparing rich integration styles versus simple linear integration styles where vision and language features are present but not integrated. After multiple comparison corrections, we tabulate the total number of electrodes that significantly pass each test as a proportion of the total number of assayed electrodes (N=1090). After aligning the location of the various electrodes to the regions provided by the Desikan-Killiany-Tourville atlas \citep{klein2012dkt}, we can further subdivide this proportion by the number of electrodes located in each region.

\section{Model Task Performance}
We give an overview of model task performance by reporting the accuracy of our 12 candidate models on several language-, vision- and multimodal-related tasks. We only include results on tasks where the model was evaluated under the same setting, e.g., zero-shot task performance. Our multimodal model performance can be seen in Table~\ref{tab:mul_performance}. The SLIP model performance and unimodal vision model performance, which is only evaluated on ImageNet can be seen in Table~\ref{tab:slip_performance}. Our unimodal language model performance is reported in Table~\ref{tab:lang_performance}. We do not report the performance the SLIP model language encoders since they have not been evaluated to our knowledge.

\begin{table}[h]
\begin{tabular}{lcccc}
\toprule
Model & VQA-v2 & NLVR$^2$ & \begin{tabular}[c]{@{}l@{}}Flickr30K (1K Test Set) Text Retrieval\\                         R@5\end{tabular} & \begin{tabular}[c]{@{}l@{}}Flickr30K Image Retrieval\\                               R@5\end{tabular} \\ \midrule
BLIP  & 77.5   & 82.3     & 99.7                                                                                                         & 96.7                                                                                                                \\
Flava & 72.5   & 78.9     & 94.0                                                                                                         & 89.38                                                                                                               \\
ALBEF & 75.8   & 83.1     & 99.5                                                                                                         & 96.3                                                                                                                \\ \bottomrule
\end{tabular}
\caption{\textbf{External Multimodal Task Performance}. Multimodal model zero-shot performance on 4 multimodal tasks. These include the VQA-v2 dataset \citep{goyal2017making}, the NLVR$^2$, dataset \citep{suhr2018corpus}, and the Flickr30K dataset \citep{plummer2015flickr30k}.}
\label{tab:mul_performance}
\end{table}

\begin{table}[h]
\begin{tabular}{lcc}
\toprule
Model       & ImageNet Linear & ImageNet Finetuned \\ \midrule
SLIP-Combo  & 72.1            & 82.6               \\
SLIP-CLIP   & 66.5            & 80.5               \\
SLIP-SimCLR & 64.0            & 82.5               \\
ConvNeXt    & 83.8            & 86.8               \\
BEIT        & 76.5            & 86.3               \\ \bottomrule
\end{tabular}
\caption{\textbf{External Unimodal Vision Task Performance}. ImageNet-1K \citep{russakovsky2015imagenet} top-1 results across the SLIP models and unimodal vision models. The left column shows performance using linear classification and the right column shows performance from finetuning.}
\label{tab:slip_performance}
\end{table}

\begin{table}[h]
\begin{tabular}{lccc}
\toprule
Model  & STS-16 & STS-B & SICK-R \\ \midrule
SBERT  & 74.3   & 77.0  & 72.9   \\
SimCSE & 80.8   & 84.2  & 80.4   \\ \bottomrule
\end{tabular}
\caption{\textbf{External Unimodal Language Model Task Performance.} Unimodal language model performance on the Semantic Task Similarity datasets \citep{pontiki2016semeval} and SICK dataset \citep{marelli-etal-2014-sick}}
\label{tab:lang_performance}
\end{table}

\begin{table}[h]
\begin{tabular}{lc}
\toprule
Model                       & Test Set Perplexity (One Movie) \\ \midrule
BLIP                        & 168.7                           \\
ALBEF                       & 223.8                           \\
Flava                       & 202.2                           \\
SLIP-Combo Language Encoder & 197.6                           \\
SLIP-CLIP Language Encoder  & 259.3                           \\
SBERT                       & 121.3                           \\
SimCSE                      & 145.4                           \\ \bottomrule
\end{tabular}
\caption{\textbf{Movie Language-based Task Performance}. Test set perplexity measured on the dialogue of a single movie across all models that have language inputs. We see that unimodal language encoders perform as well or better than multimodal encoders, which shows that our sites of multimodal integration are not associated with better unimodal processing.}
\label{tab:next-word}
\end{table}

\begin{table}[h]
\begin{tabular}{lc}
\toprule
Model                      & Scene Classification Accuracy \\ \midrule
SLIP-Combo Vision Encoder  & 73.9                              \\
SLIP-CLIP Vision Encoder   & 71.2                              \\
SLIP-SimCLR Vision Encoder & 71.0                              \\
BEIT                       & 75.4                              \\
ConvNeXt                   & 76.2                              \\
ALBEF                      & 60.1                              \\
Flava                      & 58.4                              \\
BLIP                       & 62.4                              \\ \bottomrule
\end{tabular}
\caption{\textbf{Movie Vision-based Task Performance}. Scene-Cut classification accuracy for one movie. Scene-cuts are labeled with the Places2 dataset labels and a linear classifier is trained to take image representations and decode the scene-cut label. We see that our multimodal models perform worse than unimodal vision models.}
\label{tab:scene-cut}
\end{table}

Furthermore, we report statistics on the stimulus dataset, the Brain Treebank. We train a next-word prediction task using all multimodal and language model and report the test-set perplexity in Table~\ref{tab:next-word}. We also measure the scene classification performance over the scene-cuts dataset with our multimodal and vision models and report this in Table~\ref{tab:scene-cut}.

\section{Medial Region Analysis}
\begin{figure}[h]
    \centering
    \includegraphics[width=\textwidth]{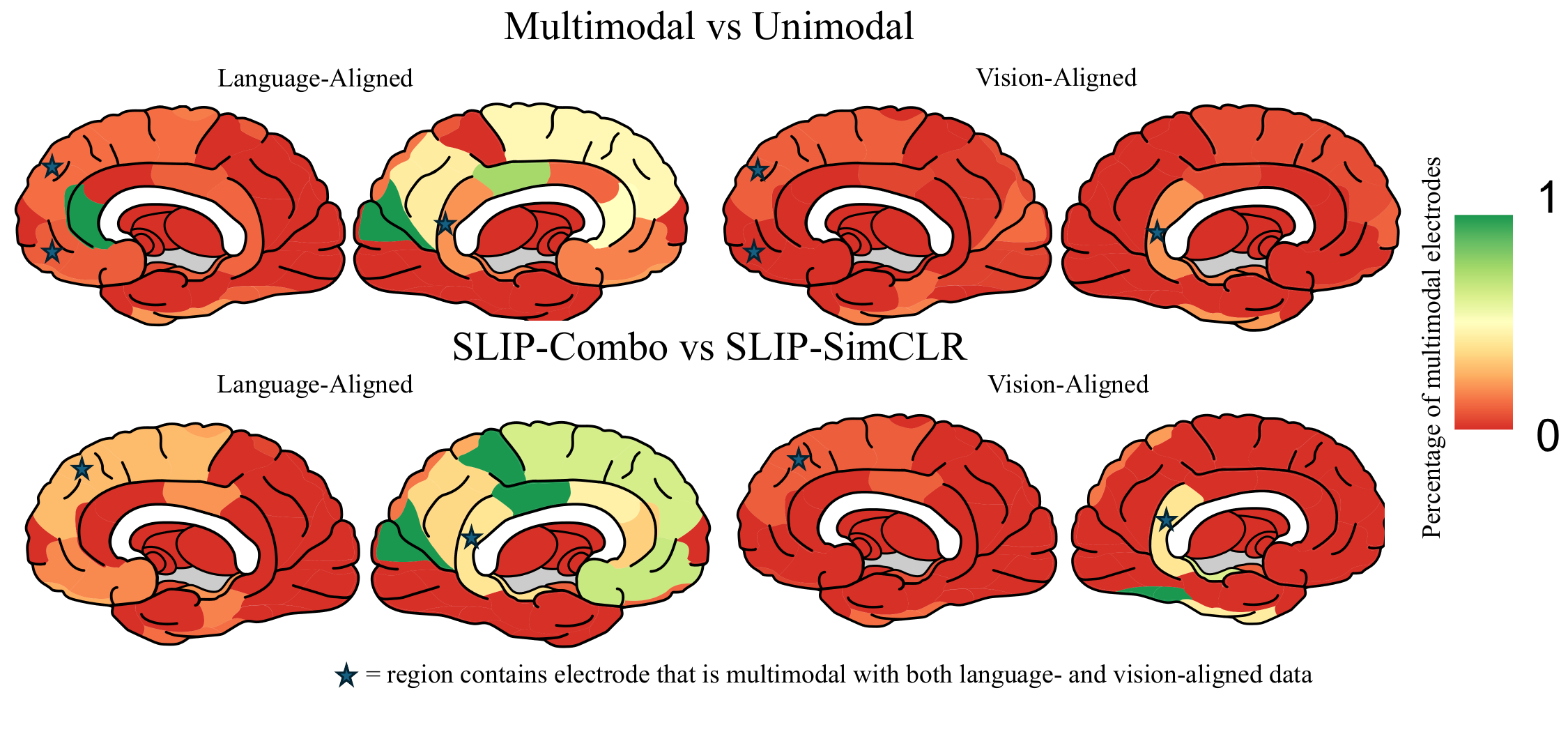}
    \caption{\textbf{Medial Region Analysis}. Multimodal sites aggregated into regions from the DKT atlas, visualizing medial regions. For each site we compute the percentage of multimodal electrodes using the first test and the (left) language or (right) vision alignment. The top defines a site to be multimodal if the best model that explains that electrode is multimodal as opposed to unimodal. The bottom controls for architecture, parameters, and datasets by comparing SLIP-Combo and SLIP-SimCLR. Red regions have no multimodal electrodes. Regions which have at least one electrode that is multimodal both with the vision and language aligned stimuli are marked with a blue star. We identify several medial regions including the superior frontal cortex and the isthmus cingulate.}
    \label{fig:mul_uni_medial}
\end{figure}
We repeat our analysis in Section~\ref{sec:mul-tests} but visualize medial regions in Figure~\ref{fig:mul_uni_medial}. We identify two main medial areas, the isthmus cingulate \citep{liu2022detection} and the superior frontal cortex \citep{willems2009differential}. This aligns with prior findings studying medial areas, identifying integration of vision, language, and action. Future work can include deeper study of medial areas but we do not carry this in our paper because we do not have many electrodes in medial regions, making results noisy.

\section{Multimodal Electrode Visualization}
\begin{figure}[h]
    \centering
    \includegraphics[width=0.8\textwidth]{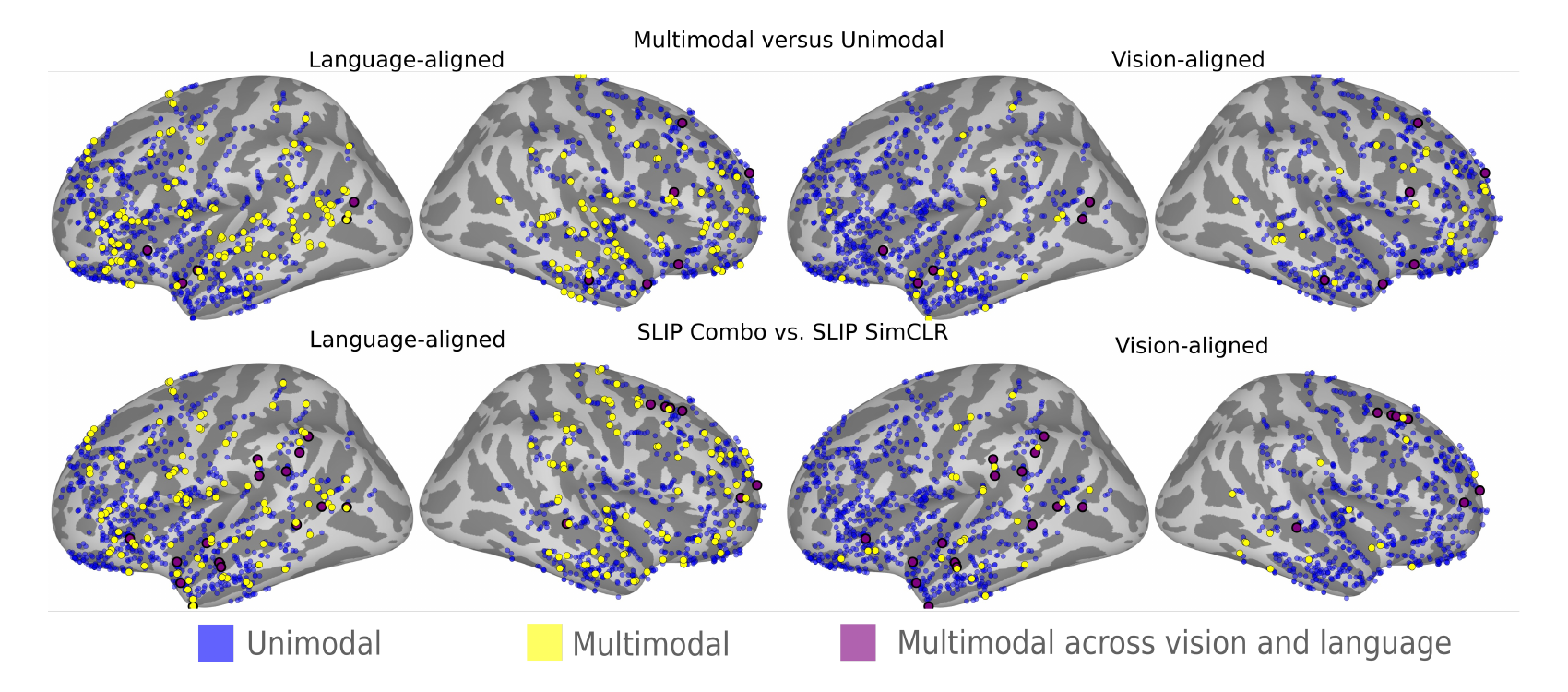}
    \caption{\textbf{Raw Multimodal Electrodes}. A raw version of Figure~\ref{fig:mul_uni} and Figure~\ref{fig:mul_uni_medial} which visualizes the electrode locations instead of aggregating over regions. We multimodal regions in a single modality in yellow and over both dataset modalities in purple.}
    \label{fig:mult-elec-viz}
\end{figure}

We show our full set of multimodal electrodes as a scatter plot in Figure~\ref{fig:mult-elec-viz}. This provides as a basis for comparison to our condensed region-based visualization.

\section{SLIP-CLIP vs SLIP-SimCLR}
\begin{figure*}
    \centering
    \includegraphics[width=0.8\textwidth]{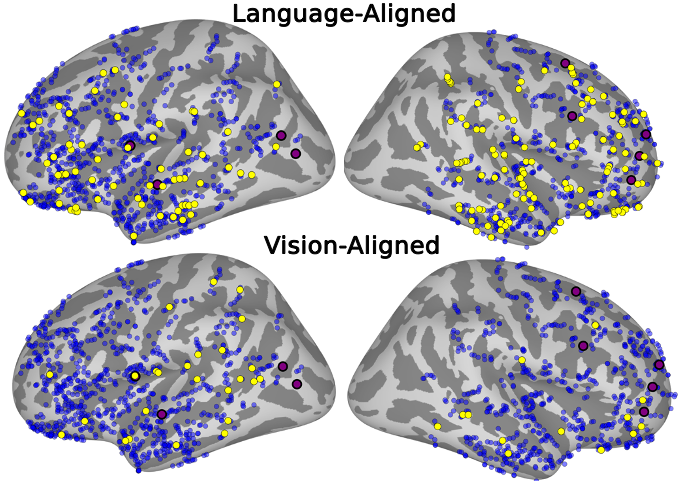}
    \caption{\textbf{SLIP-CLIP vs SLIP-SimCLR}. We compare the SLIP-CLIP vision transformer with SLIP-SimCLR instead of using SLIP-Combo. Electrodes that are better predicted by SLIP-CLIP in one dataset alignment are colored yellow. Electrodes that are better predicted by SLIP-CLIP in both dataset alignments are colored purple. We find similar electrodes are better predicted by SLIP-CLIP. A total of 9/1090 electrodes are better predicted by SLIP-CLIP than SLIP-SimCLR in both dataset alignments.}
    \label{fig:clip-simclr}
\end{figure*}

We re-run our architecture and dataset controlled multimodality tests with SLIP-CLIP and SLIP-SimCLR instead of with SLIP-Combo. We show the results in Figure~\ref{fig:clip-simclr}.

\end{document}